\documentclass{article}

\usepackage{PRIMEarxiv}

\usepackage[utf8]{inputenc} % allow utf-8 input
\usepackage[T1]{fontenc}    % use 8-bit T1 fonts
\usepackage[colorlinks=true, linkcolor=black, citecolor=black, urlcolor=black]{hyperref}
\usepackage{url}            % simple URL typesetting
\usepackage{booktabs}       % professional-quality tables
\usepackage{amsfonts}       % blackboard math symbols
\usepackage{nicefrac}       % compact symbols for 1/2, etc.
\usepackage{microtype}      % microtypography
\usepackage{lipsum}
\usepackage{fancyhdr}       % header
\usepackage{graphicx}       % graphics
\graphicspath{{media/}}     % organize your images and other figures under media/ folder

\title{

Comparing Knowledge Source Integration Methods for Optimizing Healthcare Knowledge Fusion in Rescue Operation

\thanks{\textit{2024 IEEE 7th International Conference on Industrial Cyber-Physical Systems (ICPS) | 979-8-3503-6301-2/24/\$31.00 ©2024 IEEE | DOI: 10.1109/ICPS59941.2024.10640032
}}}

\author{
  Mubaris Nadeem, Madjid Fathi \\
  Institute for Knowledge-Based Systems and Knowledge Management \\
  University of Siegen\\
  Hoelderlinstrasse 3, 57068 Siegen, Germany\\
  \texttt{\{Mubaris Nadeem\}mubaris.nadeem@uni-siegen.de} \\
}

\begin{document}
\maketitle

\begin{abstract}

In the field of medicine and healthcare, the utilization of medical expertise, based on medical knowledge combined with patients’ health information is a life-critical challenge for patients and health professionals. The within-laying complexity and variety form the need for a united approach to gather, analyze, and utilize existing knowledge of medical treatments, and medical operations to provide the ability to present knowledge for the means of accurate patient-driven decision-making. One way to achieve this is the fusion of multiple knowledge sources in healthcare. It provides health professionals the opportunity to select from multiple contextual aligned knowledge sources which enables the support for critical decisions. This paper presents multiple conceptual models for knowledge fusion in the field of medicine, based on a knowledge graph structure. It will evaluate, how knowledge fusion can be enabled and presents how to integrate various knowledge sources into the knowledge graph for rescue operations.

\end{abstract}

% keywords can be removed
\keywords{
Knowledge Fusion
\and 
Medicine
\and 
Healthcare
\and 
Integration
\and 
Knowledge Sources
\and
Knowledge Graph
\and
Bayesian Network}

\section{Introduction}

In today’s world knowledge accumulation, knowledge management, and the resulting insights are crucial for every human and system interaction, decision, and understanding. It is the foundation for complex analysis of concepts and systems. In every facet, knowledge is entrenched across all domains of the world and is no longer conceivable without it cite{rowley1999knowledge}. Especially in the field of healthcare and medicine, the significance of knowledge accumulation cannot be exaggerated cite{el2010knowledge}. In the realm of medicine, acquiring data like historical data, family medical histories, real-time vital data, and large medical data sets becomes essential for precise, personalized, and data-driven medical treatment. Hospitals has the facility of combining human expertise and the needed machinery for optimal treatment of patients in various medical specialization, however it is still needed to gather medical expertise through additional sources. In elderly care gathering various data can provide more insights in patients’ health and therefore could support medical treatment. 
\\ Knowledge can be accumulated in numerous ways and structures and can therefore lead to different insights based on the situation at hand. In the field of rescue operations artificial intelligence (AI) can, for example, use the data set to train artificial neuronal network (ANN) models to detect various diseases \cite{ ahammed_detection_2022} or can provide treatment recommendation \cite{zenkert_kirett_2022} based on expert defined treatment path to support decision-making during time-critical rescue operations. In the realm of Cyber-Physical systems (CPS), the impact of combining n-many knowledge sources is given, based on the accomplishment of a wrist-worn medical device, enabling probabilistic recommendation for rescue operations. In industry, such recommendations/extensions of knowledge can lead to improved process chains in various fields, like healthcare.

The fusion of multiple knowledge sources (e.g., medical data sets, patient records, vital signs) can be challenging and needs different approaches, based on the situation at hand. This study intends to introduce two fusion models designed for integrating knowledge sources into an established knowledge graph.

The goal of this paper is to explore and provide conceptual models for knowledge source integration in the realm of rescue operations. This paper will examine the following scientific questions:  
\begin{enumerate}
    \item How can external knowledge sources be integrated into an existing knowledge graph?
    \item How can Bayesian Network be integrated into a knowledge graph for rescue operations?
    \item Can probabilities, like Bayesian inference, help rescue operations gain more information to support their decision-making?
\end{enumerate}

With the help of knowledge fusion, it is possible to prepare larger datasets, meaning connecting various knowledge sources. In the following chapter, the state of the art in the realm of knowledge fusiona will be presented:

\begin{figure*}[t]
\includegraphics[width=\textwidth]{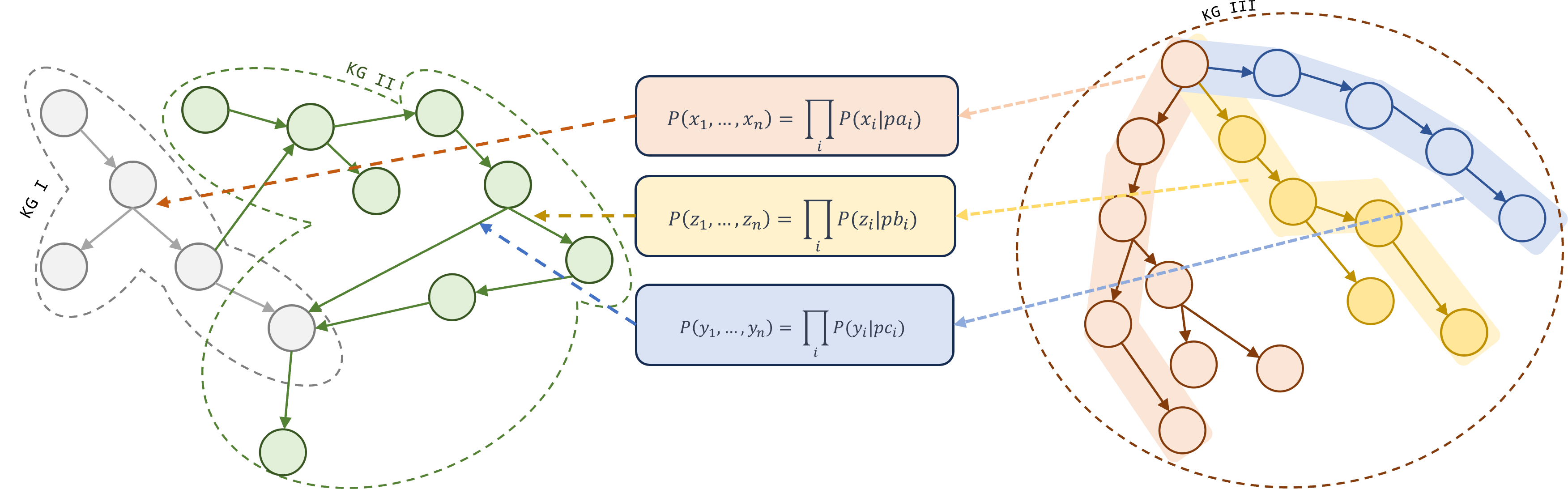}
\caption{Contextual Knowledge Fusion: \textit{KG I and KG II} presents two knowledge graphs, which are contextually interconnected with each other. \textit{KG III} presents an external knowledge source, which is based on a Bayesian network. In this, a within-calculated Bayesian interference is presented, which has an impact on the contextually aligned knowledge graphs (KG I and KG II). Such inference can be furthermore integrated on the edges of the knowledge graphs to prepare probabilities for assistive decision-making.}
\label{generalconcept}
\end{figure*}

\section{Related work}
Nguyen et al. show various versions of knowledge graph fusion in smart systems \cite{nguyen2020knowledge}. Presenting in a survey, diverse types of smart systems were identified, such as recommendation systems and question-answering systems\cite{nguyen2020knowledge}. In addition, the authors provide two approaches for knowledge graph fusion. In the first approach, knowledge of various data sources is used to construct a knowledge graph and in the second approach, Nguyen et al. integrate already existing knowledge graphs to archive one polished knowledge graph. The goal of the two approaches was to create unique knowledge graphs \cite{nguyen2020knowledge}. However, various challenges, such as real-time analysis or complex contextual relations were identified in the research from Nguyen et. al. \cite{nguyen2020knowledge}.
Zhao et al. elaborate on various tools and technologies for improved knowledge fusion, such as entity alignment and knowledge representation \cite{zhao2020multi}. The goal is to gather descriptive information from n-many sources, about the same concept, but on the focus of a data level fusion \cite{zhao2020multi}. 
Jiang and Long \cite{jiang2022knowledge} modeled a knowledge graph from various source data for a grid substation. Here the authors identified and extracted the entities and relations with named entity recognition (NER) and constructed single sources knowledge graphs, based on the individual standard. To reduce the redundancy all "single source" graphs were interconnected \cite{jiang2022knowledge}. 
All presented papers provide a deep understanding of the integration of multiple knowledge sources, using various techniques, like named entity recognition, entity alignments, and various knowledge representation methods. In the field of healthcare and medicine, such methods may support, based on the given use case, however, in treatment support, a contextual attachment of sources is needed. 
\\
The following chapter offers an exploration of the theoretical background, providing insights into the definition of knowledge, knowledge fusion, and the application of Bayesian networks (BN) and ontologies within the healthcare context.

\section{Theoretical Background}
\subsection{Knowledge}
Knowledge significantly influences individuals' life, embedding social conduct, daily tasks, or educational knowledge to provide solutions for complex tasks in various situations. This includes applied knowledge, like “cooking” or “sports” and theoretical knowledge, like “knowledge about atom physics”. Small business to global industries all bases their success on the right knowledge management \cite{yang_facilitate_2009} within their companies to achieve advantages in the international economic market. Knowledge can be differentiated into tacit \cite{nomura_knowledge_2002} and explicit knowledge \cite{yang_facilitate_2009}, where tacit knowledge refers to the type of knowledge that is embedded in individual expertise, skills, and experience. Explicit knowledge, on the other hand, can be defined as factual knowledge that can be stored and reused for and by other users \cite{yang_facilitate_2009}. \\ In the field of healthcare tacit knowledge includes the experiential knowledge of medical personnel, gained to years of experience in the field, handling medical data, critical patients and managing drugs intakes for patients. It is the knowledge which is hard to verbalize. Explicit knowledge describes medical experts-knowledge provided through scientific research and books. In the treatment of patients, experience is a basic prerequisite for correctly assessing a patient to detect, which health conditions the patient is approaching with. However, this tacit knowledge is tied to the health professional. With the SECI model, Nonaka et al. introduce knowledge conversion, which can transform knowledge from tacit to explicit and vice versa. They describe four modes of knowledge conversion: socialization, externalization, combination, and internalization. Externalization allows a conversion of tacit knowledge into explicit knowledge, which could allow to integrate years of experience into the knowledge source which then can be made accessible for a group of health professionals \cite{NONAKA20005}.

\subsection{Knowledge Sources and its Fusion}
Knowledge sources can be defined as a collection of information. It is not bound to explicit knowledge or tacit knowledge; however, its accessibility plays a significant role in the field of knowledge fusion. Using the SECI-Modell \cite{NONAKA20005} knowledge within knowledge sources can made available for the use case at hand. In the field of healthcare, knowledge sources can be field-specific, like rescue operations, medication, or precision medicine, and therefore allow better specific analysis opportunities. Knowledge, within knowledge sources, can be stored in numerous ways like in document management systems, databases in general, or in graph-based databases \cite{zenkert_kirett_2022}. Knowledge Fusion is the consolidation of multiple knowledge sources through integrating and combing information, to provide a more comprehensive understanding of the knowledge at hand.

\subsection{Knowledge Graphs}
After the introduction of Google Knowledge Graph in 2012 \cite{Zou_2020}, knowledge graphs (KG) have widely spread throughout the industrial realm. As a structure consisting of entities (describing various objects), its attributes (providing more meaningful values to the industry) and relations (the interconnection between n-manya entities) \cite{hogan}, it allows to structure knowledge and extract significant data. The data is based on the used ontologies to define semantically structured expert knowledge in the field of interest. Zou differentiates research into constructional and applications fields of knowledge graphs \cite{Zou_2020}, which differs in their focus on practical approaches in real-world scenarios. KG can be used for personalized learning recommendations \cite{hasan2023} or in the field of industrial automation \cite{liebig_building_2019}. In the realm of healthcare and medicine KG has been under investigation for various scenarios, like cancer diagnosis \cite{li_construction_2023}, clinical decision support of diabetic nephropathy \cite{lyu_causal_2023} and prediction of readmission in hospitals \cite{carvalho_knowledge_2023}.

\subsection{Ontologies in Healthcare}
Ontologies can be described as a term for allowing, sharing and reusing of knowledge bodies with technical possibilities \cite{studer1998knowledge}. They are used in knowledge-based systems for the goal of describing the domain data at hand. It consists of definitions, descriptions, and structure to support human communication, aid in interoperability, or allow inter-system communication with n-many systems \cite{zeshan_medical_2012}.
 With ontologies it is possible to structure medical terminology in a machine-readable format which helps in ensuring consistency between professionals. It allows seamless integration into other knowledge bases and supports the exchange of information throughout the system.

\subsection{Bayesian networks}
Bayesian networks (BN) are built for directed acyclic graphs, in which the entities represent the domain-specific variables and the relations represent the real-world dependencies between the entities \cite{pearl_Bayesian_2000}. In the realm of medicine and healthcare, BN have been widely used for risk estimation in medical science \cite{arora2019Bayesian}, decision-making \cite{hikal2023treatment}, or for exploration of risk factors for diseases \cite{song2023using}. It enhances the ability for health professionals to take data-driven decisions, based on numerical values and calculations, in situations, where multiple health issues cause the patients poor health \cite{mclachlan2020bayesian}. Therefore, Bayesian Network is a thought-through platform for medical settings. In its construction, BN can be differentiated into structural learning, in which the focus lies on data-driven dependencies, and parametric learning, where the weight of dependencies is integrated into the model \cite{Thanathornwong23}. With the help of Bayesian Network, it is possible to support in decision-making throughout the realm of healthcare \cite{mclachlan2020bayesian}.

\begin{figure}
\centering
\includegraphics[width=0.7\linewidth]{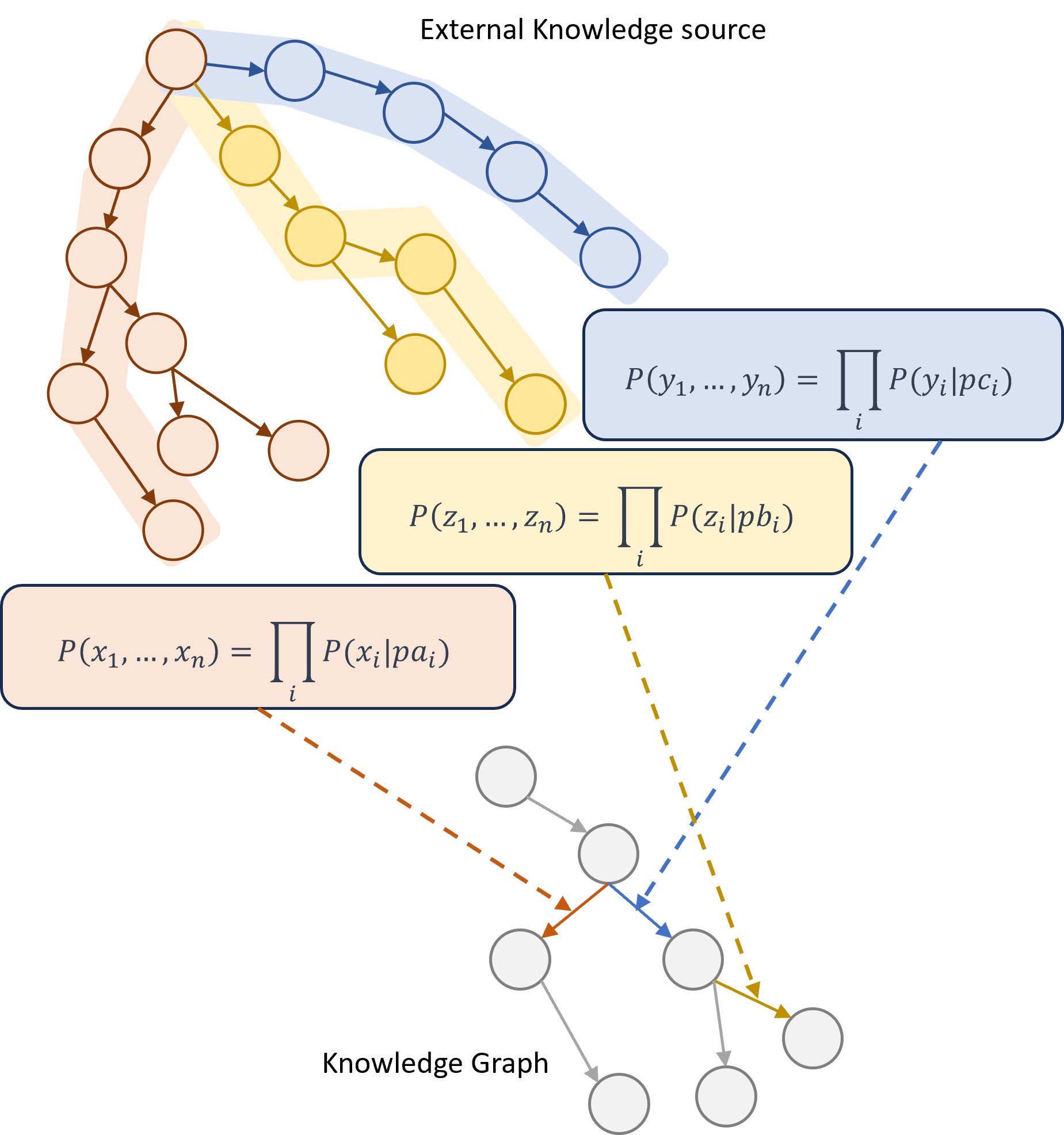}
\caption{Integration of Bayesian network into an acyclic knowledge graph. Presented are external knowledge sources as a Bayesian Network. The graph (yellow, orange, blue) describes different treatment paths, calculated inferences for each path. Those inferences are integrated as weight into the relations of the second knowledge graph, to support in decision, like treatments.  The Bayesian inference supports in decision-making for health professionals based on different external graph sources and historical data.}
\label{Bayesian}
\end{figure}

\section{Models}
This paper introduces two potential models for achieving knowledge fusion in healthcare (Fig. \ref{generalconcept}): 
\begin{itemize}
    \item Weighting relations and edges with the integration of Bayesian network.
    \item Rearrangement and Integration of n-many knowledge sources into an existing KG. 
\end{itemize} 
The following two sections will elaborate on those in more detail: 

\begin{figure*}[t]
\includegraphics[width=\textwidth]{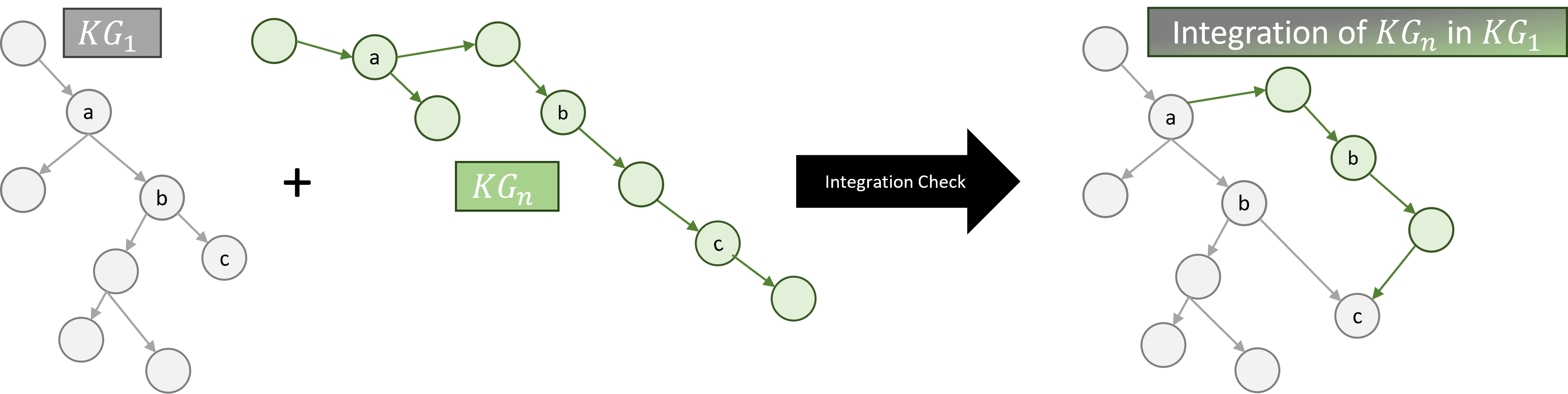}
\caption{\textbf{Knowledge fusion through contextual node-correlation:} $KG_{1}$ describes the knowledge graph, which covers a specific domain-knowledge. $KG_{n}$ describes n-many external knowledge sources (a knowledge graph, a manual, an instruction guideline) which have a contextual correlation with $KG_{1}$. The integrated KG (right) describes the resulting contextual combined knowledge graph with additional nodes for knowledge expansion on the $KG_{1}$.}
\label{integration}
\end{figure*}

\subsection{Weighting relations and edges with the integration of Bayesian network}

The general concept behind a Bayesian network infused weight integration (Fig. \ref{Bayesian}) is to support health professionals in their field (family doctor, emergencies, healthcare) with a decision-supporting algorithm that takes the domain-specific, patient-oriented, and personalized data into consideration for decision support in healthcare.  Weighting relations of the knowledge graph provide the possibility to calculate a transitional probability, which allows health professionals to identify a treatment recommendation and to evaluate whether a different treatment path is to choose.
Based on the situation at hand, health professionals then retain the ability to derive the best treatment for the current patient, while also being able to comprehend where the supporting values are derived from \cite{derks_taxonomy_2020}. In addition, the use of Bayesian networks can enable an indirect connection to external domain-specific knowledge sources. Let $ K_{infused} $ be the to be infused knowledge and let $ KG_{main} $ be the knowledge graph in which the weight integration should be integrated. Table \ref{tab1} describes some general requirements for a BN:

\begin{table}[htbp]
\caption{Requirements for Bayesian network integration in knowledge graphs}
\begin{center}
\begin{tabular}{|c|c|}
\hline
\textbf{Req.No.} & \textbf{Criteria} \\
\hline
RQ1 & $ K_{infused} $ must be from the same domain \\ \hline
RQ2 & Bayesian inference must include metadata (e.g. source) \\ \hline
RQ3 & The graph behind $ K_{infused} $ should be acyclic  \\ \hline
RQ4 & The graph behind $ K_{infused} $ should be a directed graph  \\ \hline

\hline
\end{tabular}
\label{tab1}
\end{center}
\end{table}

If those criteria can be implemented, then integration of $ P_{K_{infused}} $ into $ KG_{main} $ could be achievable. Depending on the knowledge source at hand, the probability distribution can vary, depending on the semantics of the model \cite{pearl_Bayesian_2000}. Pearl et al. differentiates between \textit{globally semantics}, in which the \textit{joint distribution} is factored into parents-based local conditional distribution,
\[ P(K_{infused}) = \prod_{i}P(K_{infused} \vert  pa_{i}) \]
and \textit{local semantics}, which states that every given variable is independent of its non-descendants \cite{pearl_Bayesian_2000}.
\[ P(K_{infused} \vert x_{1},x_{2},x_{3}) = P(K_{infused} \vert x_{2},x_{3}) \]
Based on the infused knowledge $ K_{infused} $, the probabilistic semantic needs to be decided. $x_{n}$ stands here for the descendant nodes of $K_{infused}$. The integration can be done by adding a further value to the relation/edges of $ KG_{main} $, containing the calculated Bayesian inference (Fig. \ref{Bayesian}). Further metadata, like the source of $ K_{infused} $ can be a major advantage for the user at hand. Existing KG, like PrimeKG \cite{chandak_building_2023}, typed-bio \cite{Myśliwiec_Wager_Sabat_Whiteside} or Clinical KG \cite{santos_clinical_2020} need to be analyzed if they can be used as a $ K_{infused} $ – entity. In the future, n-many KG, could be integrated into the $ KG_{main} $ and provide multiple resources for support in decision-making for health professionals. 

\subsection{Rearrangement and Integration of n-many knowledge sources into an existing KG}

The second approach, presented in this paper focuses on the knowledge integration of n-many existing knowledge graphs $ KG_{infused} $ at hand. It presents a deeper vision into the extension of KG. There are multiple ways to interconnect various knowledge sources. In \textit{PrimeKG} the authors' harmonization pipeline was based on ontology matching. They focused on the standardization of the data set format and resolved the overlap, due to similar ontologies \cite{chandak_building_2023}. This allowed them to connect 20 resources. The open-source \textit{Clinical Knowledge Graph} constructed a KG with 26 biomedical databases using 9 ontologies \cite{santos_clinical_2020}, which shows that matching databases, based on their ontology is a widely spread concept. In a scenario, where the context or environment of the treatment is important, a ontology-matching approach could help, but the context needs to be put into focus. Treatments can have similar actions (e.g. blood pressure measurement) but can differ in the results based on the context the treatment is in. This high accuracy of the right integration of sources is crucial due to the data complexity to save patient lives. Depending on the place (e.g. hospital, emergency situation) treatment equipment, drug availability, and medical expertise may differ, which could lead to a risk for the patient. In such a situation, decisions needs to be taken to provide fast and accurate medical support. 
\\A context-based knowledge fusion can provide alternative treatment paths for the situation at hand and allow health professionals to gain more knowledge out of a bigger scope of medical treatments (Fig. \ref{integration}) which enables much more accurate decision support. The figure \ref{integration} at hand shows two different knowledge graphs. The knowledge graph ($KG_{1}$) presents a treatment graph with fewer treatments of a treatment path. However the graph in $KG_{n}$ also consists of the same treatment path, but with some additional inquiries (as nodes), which suggest a more in-depth treatment. This part of the knowledge graph can then be integrated into ($KG_{1}$) to increase the broadness of the KG. With this approach, it is possible to feed the main KG (Fig. \ref{integration}, gray KG) with more and more additional graph paths, to enhance its completeness in the realm of medicine and health.

\section{Implementation}
In the following, the previously established concepts will be implemented. For that, the knowledge graph for rescue operations \cite{zenkert_kirett_2022} is used as the main $KG_{main}$ knowledge graph. 

\subsection{KIRETT}
In KIRETT, the goal of health professionals is to provide the patient in need with the required treatment for patient-stable transport to the hospital. For that, a wearable is designed to assist health professionals with a treatment recommendation. The project is differentiated between hardware and its interconnection with medical devices (e.g. Zoll X-Series), situation detection, and treatment recommendation. Initially, a situation detection, based on artificial intelligence \cite{ahammed_detection_2022} is used to detect situation groups which then leads to the specific treatment path. The treatment decision support is based on a knowledge graph for rescue operations \cite{lwda_kirett}, which was constructed through a, in practice used and expert-defined, manual \cite{treatmentpdf}. In this manual, a guideline for all rescue operation treatments is defined, which includes standard operating procedures and treatment pathways. This knowledge graph is then differentiated into various node and relation types (BPRNodes, SAANodes, $R_{n}-relations$, etc.). A user interface (UI) was designed and implemented to visualize the graph recommendation for the end user and to handle interaction with the system \cite{10187320}.

This knowledge graph will be used as the main knowledge graph $KG_{main}$ in this conceptual implementation.

\subsection{Integrating probabilities calculated by Bayesian network: An example}

\begin{figure}[htbp]
\includegraphics[width=\linewidth]{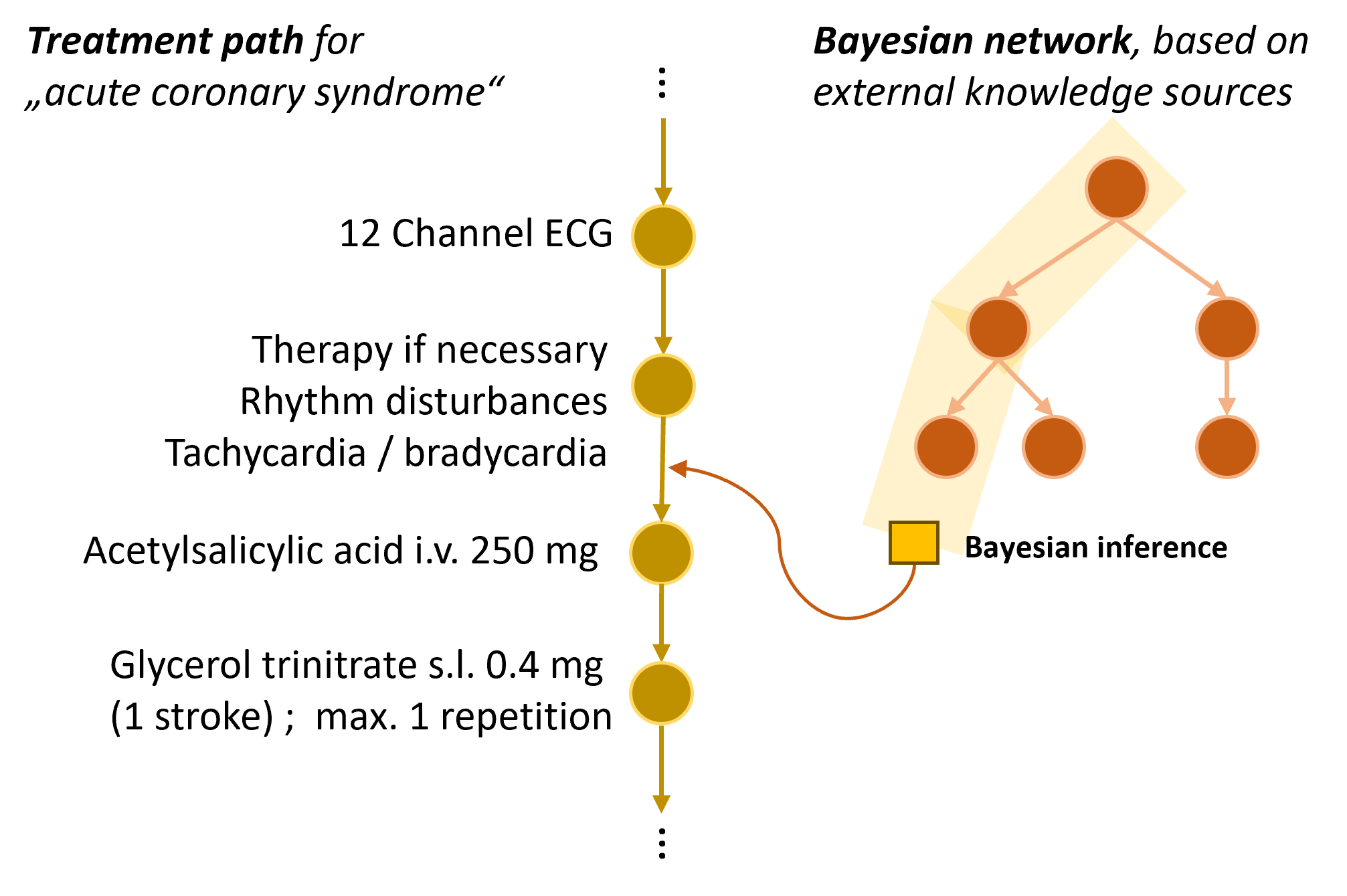}
\caption{Transfer of Bayesian inference into the knowledge graph for rescue operation. On the left side is the treatment path of the "Acute coronary syndrome" presented and on the right side the Bayesian Network of an external knowledge source is visualized. The calculated Bayesian inference can be integrated into the treatment path, to provide probabilities based on historical data for rescue operators.}
\label{bayinferance}
\end{figure}

Let $KG$ be a Bayesian network that is modeled to calculate the probability between symptoms, diseases, medication, and biomarkers. The nodes can be symptoms, like pain in the left arm or a medication like \textit{Acetylsalicylic Acid}. The relations can indicate conditional probabilities, like $P(male|Acetylsalicylic Acid)=0.7$. This could mean, that the use of acetylsalicylic acid on male patient can reduce the symptoms for the acute coronary syndrome. Constructing this as a knowledge graph can then be used to calculate the Bayesian inference for specific treatment situations. Figure \ref{bayinferance} shows a treatment path ( orange circle = treatment nodes) for the acute coronary syndrome. The Bayesian inference can now integrated to the relation to acetylsalicylic acid in the graph, allowing health professionals to reevaluate whether the medication will be helpful for the patient or not. It allows them to think about alternative, which may support a better treatment outcome. In this case the weight of the auxiliary knowledge source has no impact on the main knowledge graph itself, but rather on the decision-making process of the health professional. It introduces historical data-driven assistance for the given situation at hand. The central benefits of Bayesian Network here is, that with its help, the inference is based on the context given. It makes it easier for end user to understand the cause-and-effect of the inference, for the specific treatment.

\subsection{Knowledge Fusion through external knowledge integration}
\begin{figure}
\includegraphics[width=400px]{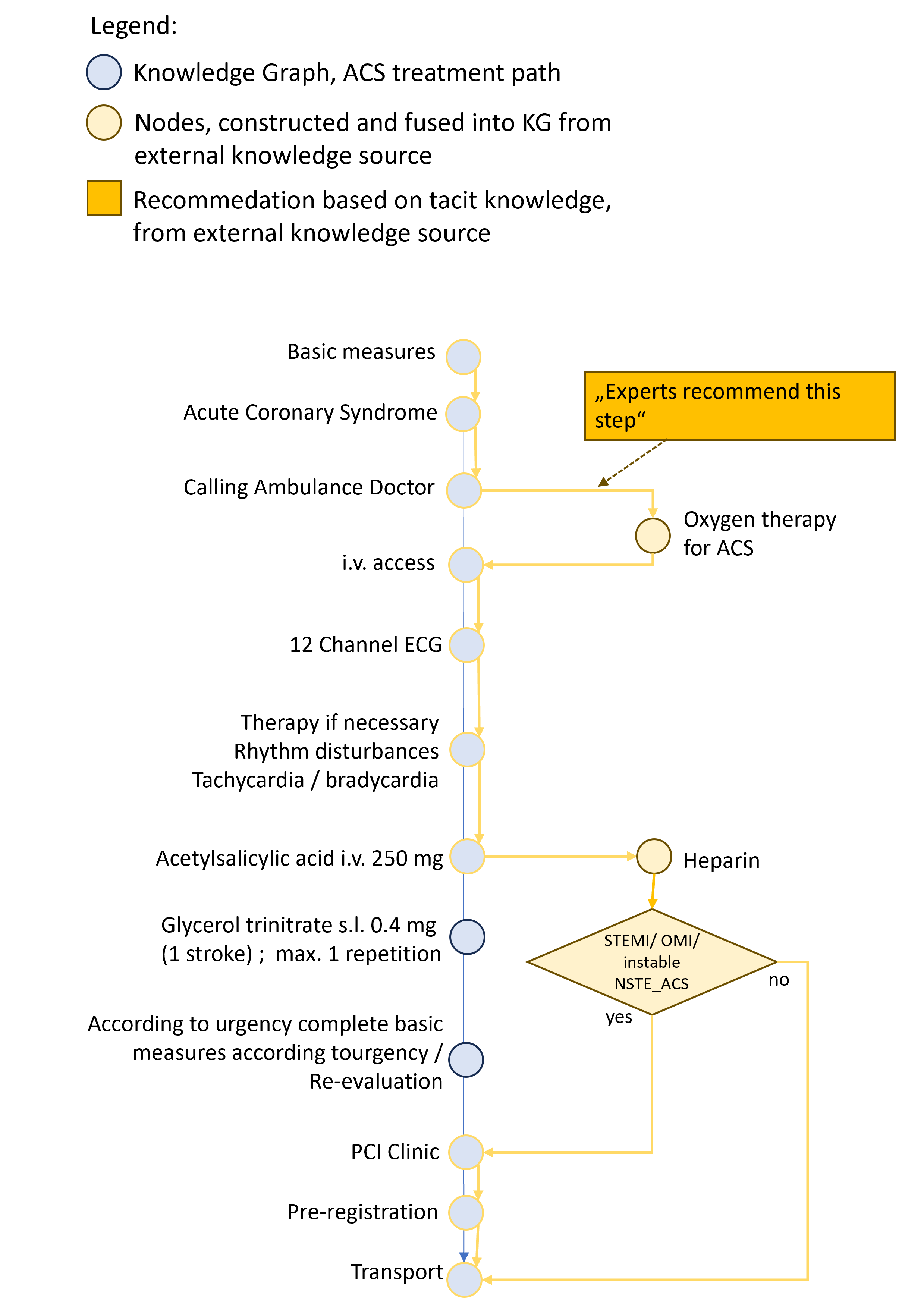}
\caption{Knowledge fusion in the treatment path of the acute coronary syndrome. The depicted flow chart is the treatment path for acute coronary syndrome, extracted from the \cite{treatmentpdf} manual for rescue operators. (Blue circle) describe the treatment path for acute coronary syndrome, constructed by medical experts \cite{treatmentpdf} and designed as a graph \cite{zenkert_kirett_2022}. (Yellow circle) presents the knowledge fusion approach based on contextual alignment. The knowledge integrated was extracted from \cite{BPR_SAA_2023}. The orange arrows describe an exemplary recommendation path, based on tacit knowledge, if external knowledge sources are integrated.}
\label{integrationacs}
\end{figure}

For knowledge fusion though external knowledge integration a knowledge graph and a knowledge source are needed:  $KG_{main}$ is defined as the KIRETT knowledge graph and the manual for rescue operations, presented by Afflerbach et al. is the defined knowledge source $K_{infused}$ \cite{BPR_SAA_2023}. $K_{infused}$, in comparison to the $KG_{main }$, also describes possible treatment path for rescue operations and uses similar ontology which provides a much more likely integration of both versions of the rescue operation. Both sources consist of the treatment path for the acute coronary syndrome. However, they differ in the position of treatment steps some additional stemps. Figure \ref{integrationacs} presents the fused knowledge graph approach. It is visible, that some additional nodes needed to be integrated, which enhanced the $KG_{main}$ with more knowledge, which may be useful for the treatment at hand. Such an integration allows health professionals to evaluate which medication would be best suitable for the patient in need. In real application, the impact differs based on the situation at hand. Drugs intake can vary and affect the patient's health differently.

\section{Discussion}

The concepts of this paper mainly describe two possible ways to integrate n-many sources of knowledge into an existing knowledge graph. The first approach, dealing with the Bayesian network aims to enable seamless integration of probabilities into a KG. The main goal hereby is to allow n-many sources to provide relevant numerical values into the graph to enrich the decision-making process by providing the Bayesian inference on relevant relations. The second approach focuses on the fusion of n-many knowledge source to extend the available knowledge for the rescue operation. Based on the contextual knowledge alignment of the graphs it provides the end user with a broader spectrum of knowledge which may lead to a better and optimized treatment. There are various ways to connect multiple data sources. \textit{PrimeKG} as an example combines various sources through identifying node types and harmonizing and extracting relationships between for example nodes\cite{chandak_building_2023}. In comparison to the presented two fusion methodologies, knowledge graphs, which are connected through node-typed identification do not consider the contextual incident at hand. The Bayesian network approach considers the context of the KG and need initial alignment with the external source to provide an accurate inference for the relation. The recommendation process here is more a decision support algorithm and less a full recommendation.  In the second, knowledge fusion concept, the contextual environment of treatment and the associated availability of medical devices, patients' medical state is considered much more. It the example presented in this paper, the acute coronary syndrome information added to the knowledge graph (Fig. \ref{integrationacs}) was possible due to the similar context of rescue operations. In summary, it is shown, that the contextual knowledge fusion, presented in this paper, can be a major advantage for healthcare and medicine.

\section{Conclusion}
The goal of this paper is to provide two conceptual models for knowledge fusion in healthcare: (a) Integration of Bayesian inference, through external knowledge sources for support in decision-making during rescue operations (b) Contextual integration of n-many knowledge sources. As foundation the knowledge graph in the project KIRETT \cite{zenkert_kirett_2022} was used. The first model is to create a relation-weighting model in knowledge graphs, which can be achieved using Bayesian networks. With the calculation of the Bayesian inference (joint distribution) it is possible to integrate further numerical information to support in rescue operations. Such inference supports health professionals, while taking actions in rescue operation, with probabilities, calculated from e.g. external sources to support in changing treatment paths. The second approach focuses on the integration of additional nodes and relations into a knowledge graph. The to be integrated knowledge should be from the same domain with contextual focus of the interoperability. Here the models focus on the contextual alignment of to-be connected graphs. Such a contextual aligned knowledge graph will extend the knowledge source and will provide health professionals to utilize more knowledge for a much-differentiated treatment.  Both models are built to enhance existing knowledge graphs with additional information, like treatments and numerical values. The values are extracted or calculated from extern knowledge sources and fused into the KG.  In this paper two conceptual implementation were shown.

\section*{Funding}
The ongoing research was financially supported by the Federal Ministry of Education and Research, Germany.  We would like to thank the associative partners of the project: Kreis Siegen-Wittgenstein, City of Siegen, the German Red Cross Siegen (DRK), and the Jung-Stilling-Hospital in Siegen.

%Bibliography
\bibliographystyle{unsrt}  
\bibliography{templateArXiv}

\end{document}